\documentclass{article}

\usepackage{PRIMEarxiv}

\usepackage{graphicx}    

\usepackage[group-separator={,}]{siunitx}
\usepackage[utf8]{inputenc} 
\usepackage[T1]{fontenc}    
\usepackage{hyperref}       
\usepackage{url}            
\usepackage{amsfonts}       
\usepackage{nicefrac}       
\usepackage{microtype}      
\usepackage{lipsum}
\usepackage{wrapfig}
\usepackage{enumitem}
\usepackage{xcolor}
\usepackage[most]{tcolorbox}
\tcbuselibrary{breakable} %
\usepackage{colortbl}
\usepackage{multirow}
\usepackage{siunitx}
\usepackage{booktabs}
\usepackage{cleveref}
\usepackage{amsmath, amsfonts, amssymb}
\usepackage[noend]{algpseudocode}
\usepackage{algorithm}
\usepackage{fancyhdr}       
\usepackage{subcaption}
\usepackage{caption} 
\usepackage{array} 
\usepackage{amsmath} 
\usepackage{rotating}
\usepackage{geometry} 
\usepackage{fontawesome5} %
\usepackage{tikz}
\usetikzlibrary{positioning,fit,shapes.geometric,shadows,arrows.meta}
\usepackage[numbers]{natbib}
\usepackage{placeins}
\usepackage{authblk}
\usepackage{academicons}
\usepackage{orcidlink}
\usepackage{xcolor}

\newcommand{\popsize}{\text{pop}_\mathrm{size}}
\newcommand{\pe}{\text{prop}_\mathrm{elite}}
\newcommand{\pmut}{\text{prop}_\mathrm{mutant}}
\newcommand{\pelite}{p_\mathrm{elite}}
\newcommand{\Tau}{\mathcal{T}}
\newcommand{\drate}{d_{\mathrm{rate}}}

\title{A Biased Random Key Genetic Algorithm for Solving the Longest Run Subsequence Problem}

\author[1]{Christian Blum~\orcidlink{0000-0002-1736-3559}}
\author[2]{Pedro Pinacho-Davidson~\orcidlink{0000-0001-9324-284X}}

\affil[1]{Artificial Intelligence Research Institute (IIIA-CSIC), Bellaterra, Spain.\protect\vspace{0.1cm} \texttt{christian.blum@iiia.csic.es}}
\affil[2]{Department of Computer Science, University of Concepcion, Concepción, Chile.\protect\vspace{0.1cm} \texttt{ppinacho@udec.cl}}

\begin{document}
\maketitle

\begin{abstract}
The longest run subsequence (LRS) problem is an NP-hard combinatorial optimization problem belonging to the class of subsequence problems from bioinformatics. In particular, the problem plays a role in genome reassembly. In this paper, we present a solution to the LRS problem using a Biased Random Key Genetic Algorithm (BRKGA). Our approach places particular focus on the computational efficiency of evaluating individuals, which involves converting vectors of gray values into valid solutions to the problem. For comparison purposes, a Max-Min Ant System is developed and implemented. This is in addition to the application of the integer linear programming solver CPLEX for solving all considered problem instances. The computation results show that the proposed BRKGA is currently a state-of-the-art technique for the LRS problem. Nevertheless, the results also show that there is room for improvement, especially in the context of input strings based on large alphabet sizes.
\end{abstract}

\keywords{Bioinformatics \and subsequence problem \and combinatorial optimization \and biased random key genetic algorithm \and experimental work}

\section{Introduction}

String and subsequence problems are fundamental in bioinformatics, where they involve analyzing biological sequences such as DNA, RNA, and proteins~\cite{erciyes2015distributed,blum2016metaheuristics}. A subsequence is a string (or sequence) derived from another string by deleting some or no elements without changing the order of the remaining elements. These problems are crucial for understanding genetic information, evolutionary relationships, and functional characteristics of biomolecules. Key string and subsequence problems include the following ones:
\begin{enumerate}
    \item \textbf{Longest Common Subsequence (LCS)}: This problem involves finding the longest string that appears as a subsequence in two or more input strings. It is used to identify conserved regions across genes or proteins, indicating functional or evolutionary significance~\cite{hirschberg1977algorithms}.
    \item \textbf{Sequence Alignment:} Subsequence analysis is integral to aligning sequences to find regions of similarity, which may suggest structural, functional, or ancestral relationships~\cite{li2010survey}.
    \item \textbf{Pattern Matching:} Finding specific motifs or subsequences within longer sequences helps identify regulatory elements, binding sites, or markers for genetic traits~\cite{de2013pattern}.
    \item \textbf{RNA Folding:} Subsequence problems assist in predicting the secondary structure of RNA by finding base-pairing regions~\cite{eddy2004rna}.
\end{enumerate}
These problems are computationally challenging due to the vast size of biological sequences, often requiring sophisticated algorithms such as dynamic programming, heuristics, and machine learning to handle the complexity and scale. \\

In this paper, we are concerned with solving a rather new member of the class of subsequence problems. The longest run subsequence (LRS) problem was first proposed in~\cite{schrinner2020longest} as a novel approach to the scaffolding phase of genome reassembly, where contigs are linked and ordered to form larger pseudo-chromosomes. This is done by means of a secondary, incomplete assembly of a related species. Specifically, the authors of~\cite{schrinner2020longest} used alignments of binned regions within a contig to identify the most homologous contig in the other assembly. The LRS problem is concerned with ordering the contigs in the secondary assembly. Apart from introducing the problem, the authors of~\cite{schrinner2020longest} proposed a dynamic programming (DP) approach for solving the problem optimally. In addition, they developed an integer linear programming (ILP) model of the problem that can be solved by a general-purpose ILP solver such as CPLEX or Gurobi. Other existing works on the LRS problem include~\cite{dondi2020longest}, in which the authors presented additional complexity results. Specifically, they demonstrated that the problem is fixed-parameter tractable when parameterized by the number of runs in a solution. Additionally, they explored the kernelization complexity of the LRS problem, proving that it does not allow for a polynomial kernel when parameterized by either the size of the alphabet or the number of runs. Lastly, they examined a restricted version of the LRS problem, where each symbol appears at most twice in the input string, and established that this variant is APX-hard. Finally, in~\cite{asahiro2023approximation}, the authors presented a polynomial-time $(k+1)/2$-approximation algorithm for the LRS problem under the $k$-occurrence constraint on input strings. For the case $k = 2$, they improved they existing approximation ratio from $(3/2)$ to $(4/3)$.

\section{Technical Problem Definition}

From a technical point of view, the LRS problem can be described as follows. An instance of the LRS problem is a tuple $(s, \Sigma)$, where $s = s_1s_2 \ldots s_{n-1}s_n$ is the so-called input string over a finite alphabet $\Sigma$. Hereby, $s_i \in \Sigma$ denotes the letter at position $i$ of $s$, and $|s| = n$ denotes the length of the input string $s$. A string $s'$ over $\Sigma$ is called a \emph{subsequence} of $s$, if it can be obtained by deleting zero or more characters from $s$. For example, given a string $s=\texttt{AGGCACT}$, the subsequence $s'=\texttt{ACCT}$ is obtained by deleting character \texttt{G} at positions 2 and 3 of $s$, and letter \texttt{A} at position 5 of $s$. A valid solution $s'$ to an LRS problem instance $(s, \Sigma)$ is any sub-sequence $s'$ of $s$ for which the following condition holds: for any pair of positions $1 \leq i < j \leq |s'|$, if ${s'}_i = {s'}_j$ (that is, if ${s'}_i$ and ${s'}_j$ are the same letters), it must hold that ${s'}_k = {s'}_i = {s'}_j$ for all $i < k < j$. In other words, the same letters appear contiguously in $s'$. Given a problem instance $(s, \Sigma)$, the goal of the LRS problem is to determine a subsequence $s^{*}$ of $s$ that holds the above-described condition and is of maximal length. For example, given the problem instance $(s=\texttt{AGGCACT}, \Sigma=\{\texttt{A}, \texttt{C}, \texttt{T}, \texttt{G}\})$, the optimal LRS solution is $s^{*} = \texttt{AGGCCT}$. In \cite{schrinner2020longest} the LRS problem was proved to be NP-hard. \\

The description of the LRS problem can be simplified based on the following observation. Every letter $s_i$ ($1 \leq i \leq n$) of a string $s = s_1 \ldots s_n$ forms part of a maximal substring $s[j,k] = s_j \ldots s_k$ of $s$, where $1 \leq j \leq i \leq k \leq n$ and $s_i = s_l$, $l=j,\ldots,k$. Such maximal substrings with the same letter at all positions are henceforth called \emph{runs}. It is not difficult to see that if an optimal solution contains the letter from a certain position of a run, it also contains the letters of all other positions of that run. Therefore, the input string $s$ can be transformed into a sequence of runs $r_1 \ldots r_m$. Hereby, each run $r_i$ represents a character $c(r_i) \in \Sigma$ and las a length $l(r_i)$. For example, the input string $s=\texttt{AGGCACT}$ can be transformed into the sequence of runs $r_1 \ldots r_6$ with:
\begin{itemize}
    \item $c(r_1) = \texttt{A}$, $l(r_1) = 1$
    \item $c(r_2) = \texttt{G}$, $l(r_2) = 2$
    \item $c(r_3) = \texttt{C}$, $l(r_3) = 1$
    \item $c(r_4) = \texttt{A}$, $l(r_4) = 1$
    \item $c(r_5) = \texttt{C}$, $l(r_5) = 1$
    \item $c(r_6) = \texttt{T}$, $l(r_6) = 1$
\end{itemize}
Let $R$ be the set that contains all these runs. A valid solution to the LRS problem can then be expressed as a subset $R'$ of $R$ such that the following holds: if $r_i, r_k \in R'$ such that $i < k$ and $c(r_i) = c(r_k)$. Then there is no $r_j \in R'$ ($i < j < k$) such that $c(r_j) \not= c(r_i)$. Among all valid solutions, the LRS problem requires finding a solution $R^{*}$ that maximizes $f(R^{*}) = \sum_{r_i \in R^{*}} l(r_i)$. Due to this simplification of the problem, a problem instance is henceforth denoted by a triple $(s, \Sigma, R)$ where $R$ is the set of runs resulting from input string $s$.

\section{Integer Linear Programming Model}

Apart from a dynamic programming approach for solving the LRS optimally, the authors of \cite{schrinner2020longest} also presented the following integer linear programming model. This model uses a binary variable $x_i$ for each run $r_i \in R$. It can be stated as follows.

\begin{align}
  \mathbf{\max}     \quad & \sum_{r_i \in R} x_i l(r_i)                         &  \label{eq:lrs:obj} \\
  \text{s.t.} && \nonumber \\
  & \sum_{\substack{i < l < j \\ c(r_i) \not= c(r_l)}} x_l \leq (2 - x_i - x_j)(j-i) &  \forall \; 1 \leq i < j \label{eq:lrs:feasibility} \\
                    & x_i \in \{0, 1\} &  \forall \; r_i \in R \nonumber 
\end{align}
Note that the objective function sums the lengths of the selected runs, while constraints \ref{eq:lrs:feasibility} makes sure that, if runs $r_i$ and $r_j$ with $c(r_i) = c(r_j)$ are selected, other runs in between $r_i$ and $r_j$ that represent letters other than $c(r_i)$ can not be selected.

\section{The Proposed Algorithm}

As mentioned earlier, we present the first metaheuristics for solving the LRS problem in this work. They are described in the following.

\subsection{BRKGA for the LRS Problem}

Our first (and principal) algorithmic approach is a Biased Random Key Genetic Algorithm (BRKGA)~\cite{gonccalves2011biased}, a metaheuristic that combines evolutionary algorithms and random key encoding principles. More specifically, the BRKGA is a Genetic Algorithm (GA) variant that uses random keys to encode solutions as real-valued vectors. These keys (values) are then mapped to problem-specific solutions. The ``biased'' part of BRKGA comes from its reproduction mechanism, which favors elite individuals during the evolution process. BRKGAs are often used to solve combinatorial optimization problems. Their main advantage is that large parts of the algorithm are problem-independent. In fact, only the translation of random keys into feasible solutions to the tackled optimization problem is problem-dependent. The problem-independent pseudo-code for the BRKGA is presented in Algorithm~\ref{alg:brkga}.

\begin{algorithm}[!t]
\caption{BRKGA pseudo-code}\label{alg:brkga}
\begin{algorithmic}[1] 
\Require a problem instance $(s, \Sigma, R)$
\Ensure values for parameters $\popsize$, $\pe$, $\pmut$, $\pelite$
    \State $P \gets \textsf{GenerateInitialPopulation}(\popsize)$
    \State \textsf{Evaluate}($P$) \Comment{problem-dependent}
    \While{\text{computation time limit not reached}}
        \State $P_e \gets \textsf{EliteSolutions}(P, \pe)$
        \State $P_m \gets \textsf{Mutants}(P, \pmut)$
        \State $P_c \gets \textsf{Crossover}(P, \pe, \pelite)$
        \State \textsf{Evaluate}($P_m \cup P_c$) \Comment{problem-dependent}
        \State $P \gets P_e \cup P_m \cup P_c$
\EndWhile
\State \Return best solution in $P$
\end{algorithmic}
\end{algorithm}

In the following, we first outline the generic or independent part of the algorithm. It begins by invoking the function \textsf{GenerateInitialPopulation}($\popsize$), which generates a population $P$ consisting of $\popsize$ randomly created individuals. Each individual $\pi \in P$ is a vector of length $m$, where $m$ denotes the number of runs of the input string. Each position in the vector of a randomly generated individual contains a random value from $[0,1]$. Subsequently, the individuals in the initial population are evaluated. Specifically, each individual $\pi \in P$ is converted into a valid solution $R_{\pi} \subseteq R$ to instance $(s, \Sigma, R)$, and the value $f'(\pi)$ for $\pi$ is defined as the objective function value $f(R_{\pi})$ of $R_{\pi}$. The process of transforming individuals into valid solutions is detailed below.

At each iteration of the algorithm, the following operations are performed. First, the best max$\{\lfloor \pe \cdot \popsize \rfloor, 1\}$ individuals are selected from $P$ and copied to $P_e$ using the function \textsf{EliteSolutions}($P, \pe$). Second, a set of max$\{\lfloor \pmut \cdot \popsize \rfloor, 1\}$ individuals, referred to as mutants, is generated and stored in $P_m$. These mutants are random individuals created in the same way as those in the initial population. Finally, a set of $\popsize - |P_e| - |P_m|$ offspring individuals is produced through crossover using the function \textsf{Crossover}($P,\pe,\pelite$) and stored in $P_c$. Each offspring individual $\pi_{\mathrm{off}} \in P_c$ is hereby generated as follows: 
\begin{enumerate}
    \item An elite parent $\pi_1$ is selected uniformly at random from $P_e$.
    \item A second parent $\pi_2$ is chosen uniformly at random from $P \setminus P_e$.
    \item For all $i = 1,\ldots,m$, the value $\pi_{\mathrm{off}}(i)$ is assigned $\pi_{1}(i)$ with probability $\pelite$, and $\pi_2(i)$ otherwise.
\end{enumerate}
Once $P_m$ and $P_c$ have been generated, these new individuals are evaluated using the function \textsf{Evaluate}(); see line 7. Remember that, in contrast to the individual from $P_m$ and $P_c$, the ones from $P_e$ are already evaluated. Finally, the population for the next generation is defined as the union of $P_e$ with $P_m$ and $P_c$.

\subsubsection{Evaluation of an Individual}
\label{sec:evaluation-individual}

The evaluation of an individual $\pi$ in function \textsf{Evaluate}() (see lines~2 and~7 of Algorithm \ref{alg:brkga}) constitutes the problem-dependent part of the proposed BRKGA algorithm. First, the $m$ runs from $R$ are ordered with respect to non-increasing grey values in $\pi$, which results in a permutation $\sigma$ of the $m$ run indices. Hereby, $\sigma(i)$ denotes the index of the run at the $i$-th position of permutation $\sigma$ ($i=1,\ldots,m$). Moreover, due to the ordering described above, it holds that $\pi(\sigma(i)) \geq \pi(\sigma(i + 1))$ for all $i=1, \ldots, m - 1$. Then, a valid solution $R_{\pi}$ is generated by traversing the runs from $r_{\sigma(1)}$ to $r_{\sigma(m)}$. At each step, $i = 1, \ldots, m$, the corresponding run $r_{\sigma(i)}$ is added to $R_{\pi}$ if and only if the validity of the resulting extended solution is preserved. 

\begin{algorithm*}
\caption{Function \textsf{Evaluate}($\pi$) of Algorithm~\ref{alg:brkga}}\label{alg:evaluate}
\begin{algorithmic}[1] 
\Require an individual $\pi$ 
\State Produce permutation $\sigma$ with respect to individual $\pi$ (see text)
\State $R_{\pi} := \emptyset$
\State $LB(a) := -1 \; \forall \; a \in \Sigma$
\State $UB(a) := -1 \; \forall \; a \in \Sigma$
\For{$i = 1, \ldots, m$}
    \State \texttt{add\_run} := \texttt{True}
    \For{all $a \in \Sigma$ s.t. $a \not= c(r_{\sigma(i)})$} \Comment{For all letters from $\Sigma$, except for the letter $c(r_{\sigma(i)})$ of the current run $r_{\sigma(i)}$}
        \If{\texttt{add\_run = \texttt{True}}}
            \State \texttt{add\_run} := \texttt{False}
            \If{$\neg(\sigma(i) > LB(a) \land \sigma(i) < UB(a))$} \Comment{Condition (1)}
                \If{$\neg(\sigma(i) < LB(a) \land LB(c(r_{\sigma(i)})) > UB(a)$} \Comment{Condition (2)}
                    \If{$\neg(\sigma(i) > UB(a) \land UB(c(r_{\sigma(i)})) < LB(a)$} \Comment{Condition (3)}
                        \State \texttt{add\_run} := \texttt{True}
                    \EndIf
                \EndIf
            \EndIf
        \EndIf
    \EndFor
    \If{\texttt{add\_run = \texttt{True}}} \Comment{The current run $r_{\sigma(i)}$ can feasibly be added to $R_{\pi}$}
        \State $R_{\pi} := R_{\pi} \cup \{r_{\sigma(i)}\}$ \Comment{$r_{\sigma(i)}$ is added to $R_{\pi}$}
        \If{$LB(c(r_{\sigma(i)})) = -1$} \Comment{If $r_{\sigma(i)}$ is the first run of $R_{\pi}$ with letter $c(r_{\sigma(i)})$}
            \State $LB(c(r_{\sigma(i)})) := \sigma(i)$
            \State $UB(c(r_{\sigma(i)})) := \sigma(i)$
        \Else
            \State \textbf{if} $\sigma(i) < LB(c(r_{\sigma(i)}))$ \textbf{then} $LB(c(r_{\sigma(i)})) := \sigma(i)$ \textbf{end if} \Comment{Update $LB(c(r_{\sigma(i)}))$ if necessary}
            \State \textbf{if} $\sigma(i) > UB(c(r_{\sigma(i)}))$ \textbf{then} $UB(c(r_{\sigma(i)})) := \sigma(i)$ \textbf{end if} \Comment{Update $UB(c(r_{\sigma(i)}))$ if necessary}
        \EndIf
    \EndIf
\EndFor
\State \Return the valid solution $R_{\pi}$
\end{algorithmic}
\end{algorithm*}

\begin{figure*}[!t]
     \centering
     \begin{subfigure}[b]{0.32\textwidth}
         \centering
         \includegraphics[width=0.99\textwidth]{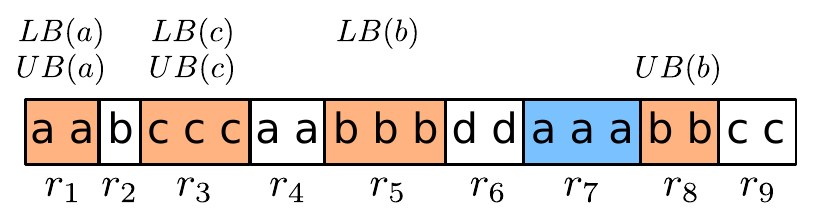}
         \caption{Example Condition 1}
         \label{fig:LRS_condition_example:1}
     \end{subfigure} 
     \begin{subfigure}[b]{0.32\textwidth}
         \centering
         \includegraphics[width=0.99\textwidth]{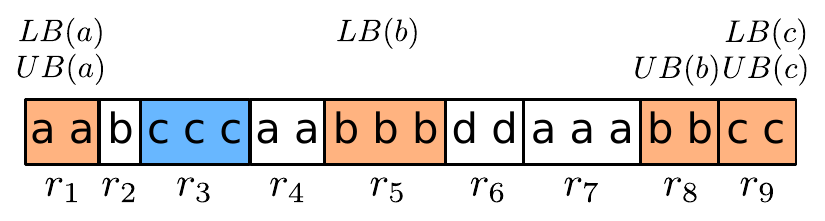}
         \caption{Example Condition 2}
         \label{fig:LRS_condition_example:2}
     \end{subfigure} 
     \begin{subfigure}[b]{0.32\textwidth}
         \centering
         \includegraphics[width=0.99\textwidth]{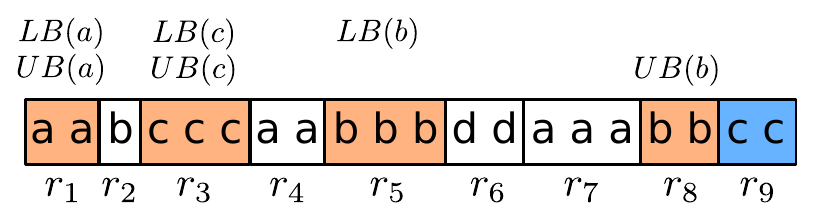}
         \caption{Example Condition 3}
         \label{fig:LRS_condition_example:3}
     \end{subfigure}
        \caption{The examples shown in this graphic consider the input string $s =$\texttt{aabcccaabbbddaaabbcc} over the alphabet $\Sigma = \{\texttt{a}, \texttt{b}, \texttt{c}, \texttt{d}\}$. The three graphics show that this input string results in 9 runs of different lengths $R = \{r_1, \ldots, r_9\}$. The runs already added to the solution under construction are indicated in orange in the three examples. In contrast, the run that is checked if it can feasibly be added is indicated in blue color.  Moreover, the $LB()$ and $UB()$ values of the letters are indicated by their position above the input string. (a) shows an example for condition 1. Run $r_7$ cannot be added to the solution because $LB(\texttt{b}) < 7 < UB(\texttt{b})$. (b) shows an example for condition 2. Run $r_3$ (with $c(r_3) =$ \texttt{c} cannot be added to the solution because $3 < LB(\texttt{b})$ and $UB(\texttt{b}) < LB(\texttt{c})$. Finally, (c) shows an example for condition 3. Run $r_9$ (with $c(r_9) =$\texttt{c} cannot be added to the solution because $9 > UB(\texttt{b})$ and $UB(\texttt{c}) < LB(\texttt{b})$.}
        \label{fig:LRS_condition_example}
\end{figure*}

Note that the efficient implementation of evaluating an individual $\pi$ is crucial for the algorithm's success. In fact, we have put a lot of emphasis on this aspect. The most efficient implementation that we could find is pseudo-coded in Algorithm~\ref{alg:evaluate}. This implementation works with two data structures, $LB()$ and $UB()$, henceforth called the letter lower and upper bounds. $LB(a)$ stores for each letter $a \in \Sigma$ the smallest index of all runs that are added to solution $R_{\pi}$. Similarly, $UB(a)$ stores for each letter $a \in \Sigma$ the largest index of all runs that are added to solution $R_{\pi}$. Both data structures are (for all $a \in \Sigma$) initialized to -1 (see lines~3 and~4 of Algorithm~\ref{alg:evaluate}). Then, for all $i=1,\ldots,m$, the corresponding run $r_{\sigma(i)}$ is added to $R_{\pi}$ only if for all letters $\{a \in \Sigma \mid a \not= c(r_{\sigma(i)})$ the following three conditions are fulfilled (see also lines~10 to~12 of Algorithm~\ref{alg:evaluate}):
\begin{itemize}
    \item \textbf{Condition 1:} $\neg(\sigma(i) > LB(a) \land \sigma(i) < UB(a))$, that is, the index $\sigma(i)$ of the run that is currently evaluated should not be in between the smallest and the largest index of runs of any letter $a \not = c(r_{\sigma(i)})$; see Figure~\ref{fig:LRS_condition_example:1} for an example.
    \item \textbf{Condition 2:} $\neg(\sigma(i) < LB(a) \land LB(c(r_{\sigma(i)})) > UB(a))$, that is, the index $\sigma(i)$ of the run that is currently evaluated should not be smaller than $LB(a)$ for some letter $a \in \Sigma$, while $LB(c(r_{\sigma(i)}))$ (the $LB()$ of the letter of the run currently under evaluation) is greater than $UB(a)$; see Figure~\ref{fig:LRS_condition_example:2} for an example.
    \item \textbf{Condition 3:} $\neg(\sigma(i) > UB(a) \land UB(c(r_{\sigma(i)})) < LB(a))$, that is, the index $\sigma(i)$ of the run that is currently evaluated should not be greater than $UB(a)$ for some letter $a \in \Sigma$, while $UB(c(r_{\sigma(i)}))$ (the $UB()$ of the letter of the run currently under evaluation) is smaller than $LB(a)$; see Figure~\ref{fig:LRS_condition_example:3} for an example.
\end{itemize}

If, after checking all conditions, the run under evaluation can be added to the solution under construction, the $LB()$ and $UB()$ values of the letter of the added run must be updated as shown in lines~21 to~27 of Algorithm~\ref{alg:evaluate}.

\subsection{ACO for the LRS Problem}

As mentioned in the introduction, to compare it with the previously introduced BRKGA approach, we also implemented a Max-Min Ant System algorithm in the hypercube framework~\cite{blum2004hyper}, henceforth simply called ACO. This algorithm is based on the principles of the standard Max-Min Ant System~\cite{stutzle2000max}. However, the pheromone values are confined to the pre-defined range $[0,1]$. By limiting the pheromone values to such a hypercube structure, the algorithm ensures controlled exploration and exploitation, enhancing stability and performance on complex optimization problems.

In particular, we used the variant of the Max-Min Ant System algorithm for subset selection problems exactly as outlined, for example, in~\cite{nurcahyadi2021adding}. This algorithm variant is problem-independent, except for the solution construction procedure. It uses a set $\Tau$ of $m$ pheromone values, where $m$ is again the number of total runs, $|R|$, that are present in the given problem instance. More specifically, $\Tau$ contains for each run $r_i \in R$ a pheromone value $0 \leq \tau_i \leq 1$. \\

Constructing valid LRS solutions based on the pheromone values is actually very similar to the \textsf{Evaluate}() function of the BRKGA algorithm from Algorithm~\ref{alg:evaluate}. Only the way of generating a permutation $\sigma$ of the run indices is different. While, in BRKGA, $\sigma$ is obtained on the basis of the grey values of the runs in the individual $\pi$ to be evaluated, in ACO $\sigma$ is produced as shown in Algorithm~\ref{alg:sigma-generation}. As can be seen, at each step of generation $\sigma$, with probability $\drate$ a run index $j$ for position $i$ of $\sigma$ is chosen deterministically as the one with maximal value $\tau_k \in l(r_k)$ for all $k \in I$. In this context, remember that $l(r_k)$ denotes the length of run $r_k$. Otherwise, with probability $1 - \drate$, the run index $j$ for position $i$ of $\sigma$ is selected by roulette-wheel selection based on the above-mentioned values.

\begin{algorithm}
\caption{Pheromone-based generation of permutation $\sigma$}\label{alg:sigma-generation}
\begin{algorithmic}[1] 
\Require the set of pheromone values $\Tau$, a determinism rate $0 \leq \drate \leq 1$ 
\State Let $I = \{1, \ldots, m\}$ be the set of all run indices
\For{$i=1, \ldots, m$}
    \State Draw a random number $r \in [0,1]$ uniformly at random
    \If{$r \leq \drate$}
        \State $j := \text{argmax}\{ \tau_k \cdot l(r_k) \mid k \in I\}$
    \Else
        \State Choose a run index $j \in I$ with roulette-wheel selection based on values $\tau_k \cdot l(r_k)$ for all $k \in I$
    \EndIf
    \State $\sigma(i) := j$
    \State $I := I \setminus \{j\}$
\EndFor
\State \Return permutation $\sigma$
\end{algorithmic}
\end{algorithm}

\section{Experimental Evaluation}
\label{sec:experiments}

Both BRKGA and ACO were implemented in C++ and compared to the application of the ILP solver CPLEX (version 22.1). Hereby, CPLEX is used in sequential mode in order to ensure a fair comparison. For conducting the experiments, we used a high-performance computing cluster of machines equipped with Intel\textsuperscript{\textregistered} Xeon\textsuperscript{\textregistered} 5670 CPUs having 12 cores of 2.933 GHz and at least 32~GB of RAM.

\subsection{Benchmark Instances}

In the absence of a suitable benchmark set, a collection of artificial instances was generated. Specifically, for each combination of the input string length $n \in {100, 200, 300, 500, 1000, 2000, 5000}$ and the alphabet size $|\Sigma| \in \{2, 4, 8, 16, 32\}$, 30 problem instances were generated uniformly at random. This makes a total of 1050 problem instances. They can be obtained from the corresponding author on request.

\subsection{Algorithm Tuning}

Both BRKGA and ACO require carefully chosen parameter settings to perform at their best. To achieve this, we utilized \texttt{irace}\footnote{\url{https://mlopez-ibanez.github.io/irace/}}, a scientific tool for parameter optimization. The tuning process for both BRKGA and ACO was conducted once across the entire benchmark set. To facilitate this, additional problem instances were generated, with one tuning instance created for each combination of $n$ and $|\Sigma|$, resulting in a total of 35 instances. Each algorithm was allocated a tuning budget of 2000 runs, with computation time limits defined based on the input string length as $n$/10 CPU seconds. \\

Even though lacking the space to describe the ACO parameters, we nevertheless provide their final settings: 
\begin{itemize}
    \item Solution constructions per algorithm iteration: 10
    \item Learning rate (pheromone decay): $0.33$
    \item Determinism rate ($\drate$): 0.92
\end{itemize}
We refer to~\cite{nurcahyadi2021adding} for an explanation regarding these parameters. \\

\begin{table}[!t]
    \centering
    \caption{BRKGA tuning results}
    \label{tab:tuning_results}
    \begin{tabular}{l|l|r} \toprule
    \textbf{Parameter} & \textbf{Domain} & \textbf{Chosen Value}   \\ \midrule
    $\popsize$    & $\{10, \ldots, 500\}$                & 356              \\
    $\pe$         & $[0.1, 0.25]$                       & 0.18            \\
    $\pmut$       & $[0.1, 0.3]$                        & 0.29            \\
    $\pelite$     & $[0.51, 0.8]$                       & 0.69            \\
    \bottomrule
    \end{tabular}
\end{table}

The tuning results for BRKGA, together with the allowed parameter domains, are shown in Table~\ref{tab:tuning_results}. Interestingly, the population size (356) is rather large. This helps the algorithm presumably to avoid getting stuck in local minima. Note that such large population sizes can only be afforded due to the efficient implementation of the process of evaluating an individual (see Section~\ref{sec:evaluation-individual}). 

\subsection{Results}

\begin{table*}[!t]
    \centering
    \caption{Numerical Results}
    \label{tab:results}
    \begin{tabular}{rr|rrr|rr|rr} \toprule
    && \multicolumn{3}{c|}{\textbf{CPLEX}} & \multicolumn{2}{c|}{\textbf{ACO}}  & \multicolumn{2}{c}{\textbf{BRKGA}} \\ \cmidrule{3-5} \cmidrule{6-7} \cmidrule{8-9}
    Length & $|\Sigma|$ & average & time & gap & average & time & average & time \\ \midrule
     &  2  &  \bf59.03  &  0.11  &  8.55         &  \bf59.03  &  0.01  &  \bf59.03  &  0.00 \\
     &  4  &  40.97  &  1.00  &  44.16           &  40.70  &  1.35  &  \bf41.07  &  0.05 \\
100  &  8  &  33.33  &  4.57  &  64.98           &  30.57  &  3.04  &  \bf33.67  &  0.52 \\
     &  16  &  \bf34.20  &  6.46  &  57.59       &  27.63  &  2.62  &  33.77  &  2.45 \\
     &  32  &  \bf42.03  &  4.76  &  33.13       &  28.73  &  3.30  &  39.53  &  4.65 \\ \midrule
     &  2  &  \bf115.17  &  1.19  &  11.22       &  \bf115.17  &  0.14  &  \bf115.17  &  0.00 \\
     &  4  &  70.63  &  8.22  &  77.08           &  71.77  &  4.56  &  \bf72.77  &  0.12 \\
200  &  8  &  51.83  &  14.12  &  135.37         &  49.27  &  8.93  &  \bf55.03  &  2.52 \\
     &  16  &  46.00  &  12.59  &  165.92        &  35.90  &  6.48  &  \bf49.97  &  6.06 \\
     &  32  &  \bf53.90  &  15.96  &  123.77     &  27.87  &  7.16  &  52.90  &  11.25 \\ \midrule
     &  2  &  \bf165.47  &  8.37  &  12.54       &  \bf165.47  &  0.10  &  \bf165.47  &  0.00 \\
     &  4  &  95.77  &  9.82  &  103.99          &  101.53  &  5.54  &  \bf102.63  &  0.40 \\
300  &  8  &  64.43  &  13.85  &  204.36         &  66.53  &  15.56  &  \bf74.63  &  3.72 \\
     &  16  &  55.07  &  13.99  &  253.98        &  50.00  &  11.56  &  \bf65.23  &  10.00 \\
     &  32  &  61.03  &  21.54  &  211.73        &  45.33  &  9.73  &  \bf63.73  &  19.81 \\ \midrule
     &  2  &  269.27  &  14.76  &  21.87         &  \bf271.67  &  0.42  &  \bf271.67  &  0.02 \\
     &  4  &  146.73  &  27.21  &  125.02        &  161.00  &  7.04  &  \bf162.73  &  1.44 \\
500  &  8  &  84.87  &  28.42  &  292.47         &  99.97  &  21.77  &  \bf109.70  &  7.34 \\
     &  16  &  61.63  &  25.91  &  441.28        &  66.93  &  21.72  &  \bf88.63  &  23.13 \\
     &  32  &  66.03  &  32.57  &  401.36        &  57.33  &  17.16  &  \bf81.43  &  31.73 \\ \midrule
      &  2  &  458.90  &  43.60  &  $>$10000.00  &  \bf530.23  &  0.85  &  \bf530.23  &  0.03 \\
      &  4  &  196.57  &  50.00  &  $>$10000.00  &  297.73  &  35.99  &  \bf300.83  &  9.97 \\
1000  &  8  &  85.73  &  40.95  &  $>$10000.00   &  175.73  &  56.10  &  \bf189.93  &  20.81 \\
      &  16  &  49.73  &  28.08  &  $>$10000.00  &  105.63  &  41.94  &  \bf139.60  &  47.00 \\
      &  32  &  42.67  &  23.21  &  $>$10000.00  &  67.90  &  41.13  &  \bf119.20  &  71.98 \\ \midrule
      &  2  &  0.00  &  200.00  &  $>$10000.00   &  \bf1041.73  &  16.07  &  \bf1041.73  &  1.53 \\
      &  4  &  0.00  &  198.53  &  $>$10000.00   &  567.57  &  74.70  &  \bf572.10  &  5.88 \\
2000  &  8  &  0.00  &  178.90  &  $>$10000.00   &  317.13  &  99.79  &  \bf342.13  &  55.61 \\
      &  16  &  0.00  &  106.48  &  $>$10000.00  &  180.77  &  103.18  &  \bf230.80  &  107.27 \\
      &  32  &  0.00  &  52.91  &  $>$10000.00   &  115.43  &  83.32  &  \bf177.47  &  154.92 \\ \midrule
      &  2  &  0.00  &  500.00  &  $>$10000.00   &  2567.37  &  41.68  &  \bf2567.47  &  1.14 \\
      &  4  &  0.00  &  500.00  &  $>$10000.00   &  1350.33  &  274.30  &  \bf1360.47  &  24.65 \\
5000  &  8  &  0.00  &  500.00  &  $>$10000.00   &  720.67  &  266.67  &  \bf763.17  &  221.00 \\
      &  16  &  0.00  &  500.00  &  $>$10000.00  &  394.30  &  301.26  &  \bf476.20  &  343.23 \\
      &  32  &  0.00  &  500.00  &  $>$10000.00  &  220.67  &  223.14  &  \bf322.00  &  419.01 \\ \bottomrule
\end{tabular}
\end{table*}

Each of the three algorithmic techniques (BRKGA, ACO, and CPLEX) was applied once to every problem instance of the benchmark set, using the same computation time limits specified during the description of parameter tuning (see the previous section). The results are presented in Table~\ref{tab:results}. Every row in that table shows the results averaged over the 30 problem instances of the corresponding combination of $n$ (string length) and $|\Sigma|$ (alphabet size). For all three algorithms, the results are presented in terms of the average solution quality (calculated over 30 problem instances) in the column labeled ``average,'' and the average time (in CPU seconds) required to find these solutions in the column labeled ``time.'' In the case of CPLEX, a third column provides information about the optimality gap at the end of a run.\footnote{A gap of zero would indicate that CPLEX found (provenly) optimal solutions in all 30 cases. However, this does not even happen in the case of the smallest problem instances of string length $n=100$.} Finally, note that the best result in each table row is marked in bold font. \\

\begin{figure}[!t]
     \centering
     \begin{subfigure}[b]{0.4\textwidth}
         \centering
         \includegraphics[width=0.95\textwidth]{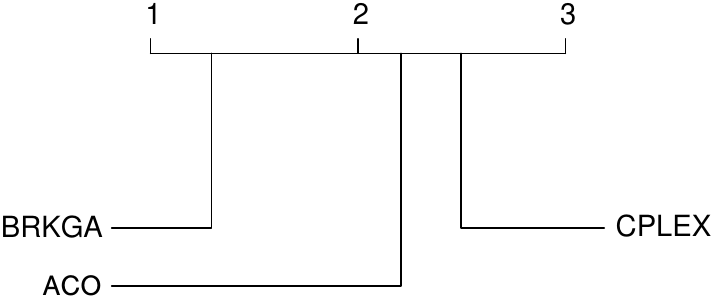}
         \caption{All instances}
         \label{fig:CD-plots:1}
     \end{subfigure} 
     \begin{subfigure}[b]{0.4\textwidth}
         \centering
         \includegraphics[width=0.95\textwidth]{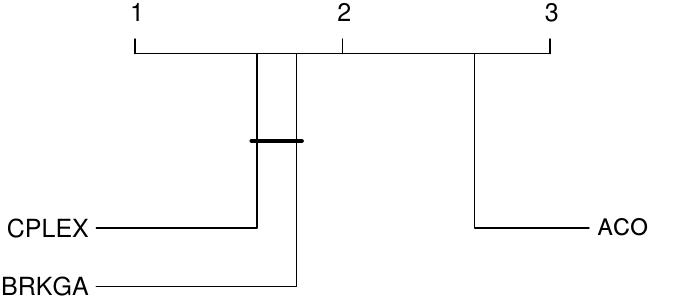}
         \caption{Instances with $n=100$}
         \label{fig:CD-plots:2}
     \end{subfigure} 
        \caption{Critical difference (CD) plots allowing for a statistical evaluation of the results}
        \label{fig:CD-plots}
\end{figure}

To support the analysis of the results with statistical significance claims, so-called critical difference (CD) plots---produced with the R package \texttt{scmamp}\footnote{\url{https://github.com/b0rxa/scmamp}}---are provided in Figure~\ref{fig:CD-plots}. This package offers functions for generating CD plots based on the outcomes of statistical comparisons. Vertical whiskers in these plots represent the average rankings of the algorithms. Bold horizontal bars connecting the whiskers indicate that the performance differences between the corresponding algorithms are not statistically significant. Conversely, the absence of a bold horizontal bar between two whiskers implies a statistically significant difference in their performance. In particular, Figure~\ref{fig:CD-plots} provides a CD plot concerning all problem instances (a), in addition to a CD plot that considers only the shortest input strings of $n=100$ (b). \\

\noindent The results allow us to make the following observations:
\begin{enumerate}
    \item First of all, BRKGA is clearly the best-performing algorithm. Considering the whole benchmark set, BRKGA outperforms the two competitors (ACO and CPLEX) with statistical significance; see Figure~\ref{fig:CD-plots:1}.
    \item However, the results for the shortest input strings ($n=100$) indicate that, with growing alphabet size, BRKGA does not seem to find optimal solutions. Note that, in the cases ($n=100$, $|\Sigma| \in \{16, 32\}$), CPLEX shows a better performance than BRKGA. This is also confirmed by the CD plot that only considers instances with $n=100$ in Figure~\ref{fig:CD-plots:2}.
    \item Concerning the comparison between BRKGA and ACO, we can state that ACO is only able to perform at the same level as BRKGA for problem instances with the smallest possible alphabet size of $|\Sigma|=2$. However, note that---in these cases---ACO requires much more computation time to obtain comparable results. With growing alphabet size, ACO shows an increasing disadvantage with respect to BRKGA.
\end{enumerate}

\begin{figure}
    \centering
    \includegraphics[width=0.4\textwidth]{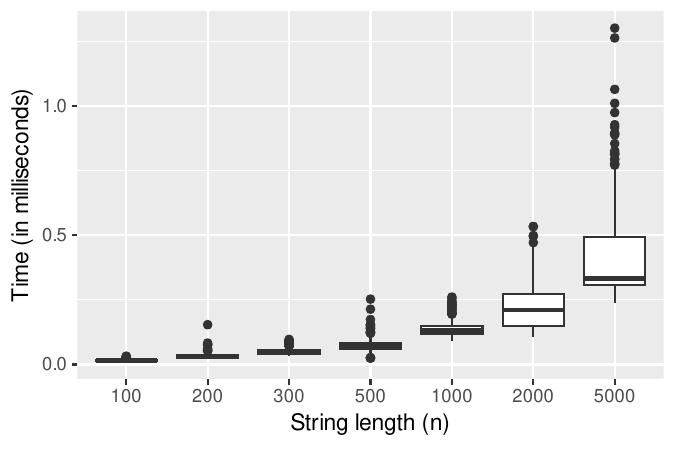}
    \caption{Solution construction times (in milliseconds)}
    \label{fig:constr_times}
\end{figure}

\begin{figure*}[!h]
     \centering
     \begin{subfigure}[b]{0.49\textwidth}
         \centering
         \includegraphics[width=0.85\textwidth]{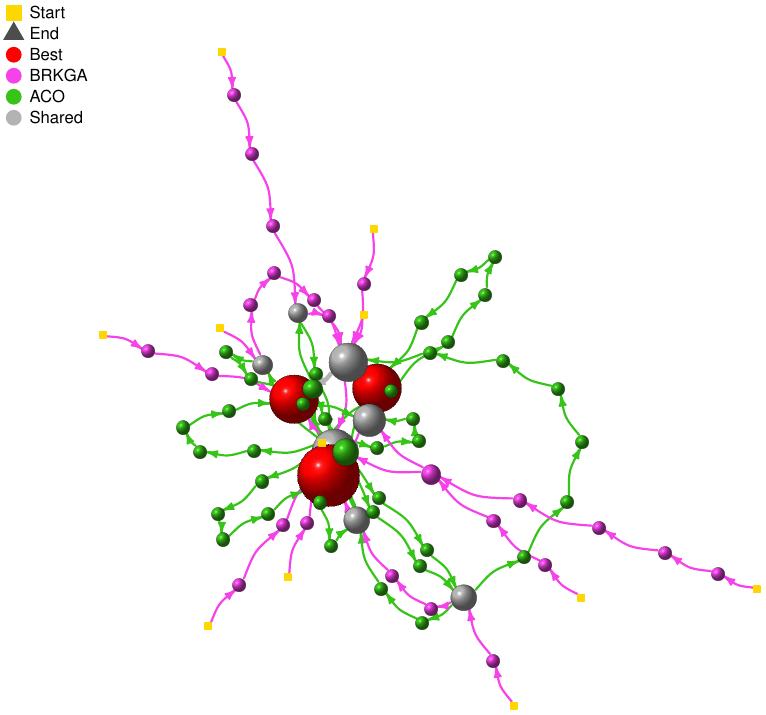}
         \caption{$n=5000$, $|\Sigma|=2$}
         \label{fig:STNWeb-graphics:1}
     \end{subfigure} \\
     \begin{subfigure}[b]{0.49\textwidth}
         \centering
         \includegraphics[width=0.85\textwidth]{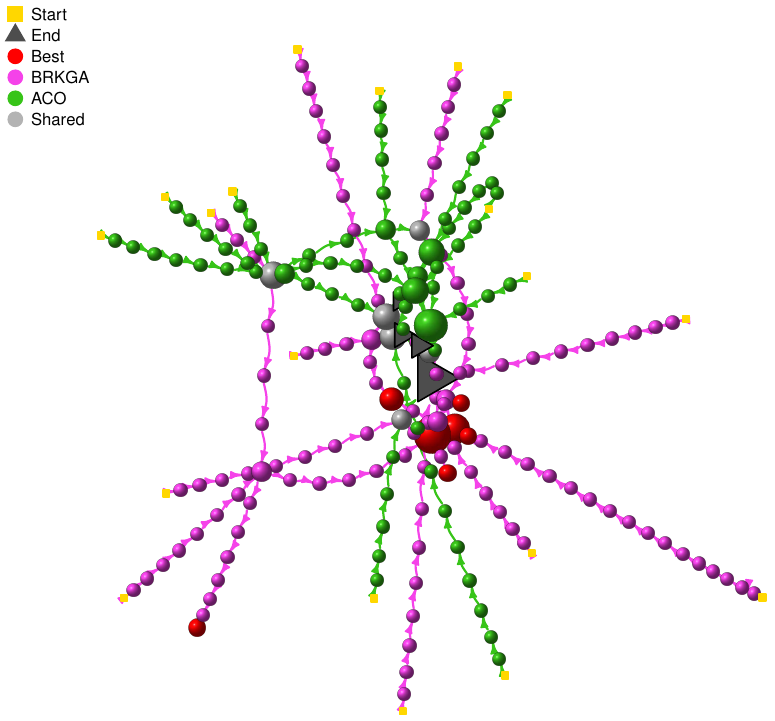}
         \caption{$n=5000$, $|\Sigma|=4$}
         \label{fig:STNWeb-graphics:2}
     \end{subfigure} 
     \begin{subfigure}[b]{0.49\textwidth}
         \centering
         \includegraphics[width=0.85\textwidth]{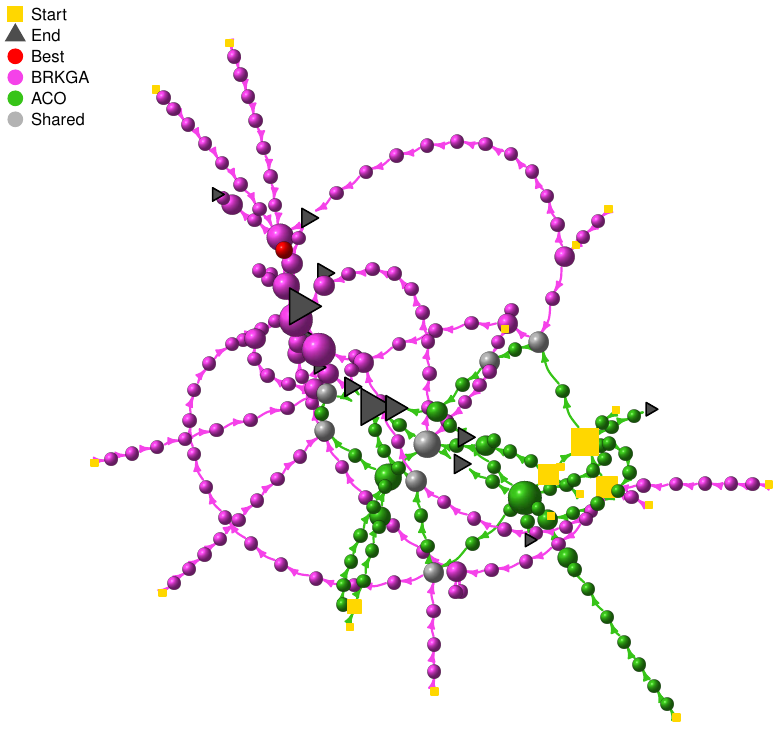}
         \caption{$n=5000$, $|\Sigma|=8$}
         \label{fig:STNWeb-graphics:3}
     \end{subfigure} \\
     \begin{subfigure}[b]{0.49\textwidth}
         \centering
         \includegraphics[width=0.85\textwidth]{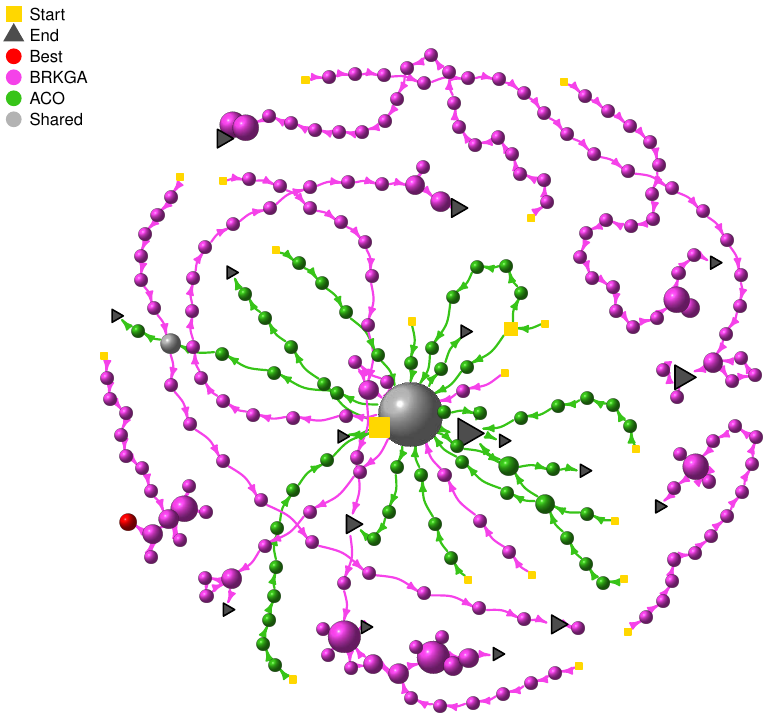}
         \caption{$n=5000$, $|\Sigma|=16$}
         \label{fig:STNWeb-graphics:4}
     \end{subfigure} 
     \begin{subfigure}[b]{0.49\textwidth}
         \centering
         \includegraphics[width=0.85\textwidth]{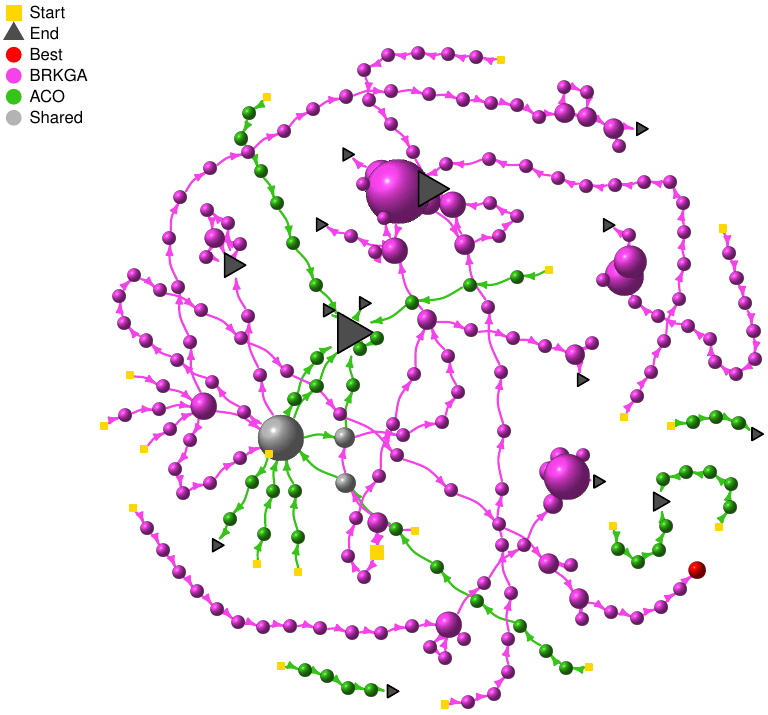}
         \caption{$n=5000$, $|\Sigma|=32$}
         \label{fig:STNWeb-graphics:5}
     \end{subfigure} 

        \caption{STNWeb graphics. The five graphics show 10 runs of BRKGA and ACO for the first (out of 30) problem instances with $n=5000$ and $|\Sigma| \in \{2, 4, 8, 16, 32\}$.}
        \label{fig:STNWeb-graphics}
\end{figure*}

To demonstrate the efficiency of the implementation for evaluating individuals in the BRKGA algorithm, we measured the time required to evaluate each of 100 random solutions to the first problem instance of each combination of $n$ and $|\Sigma|$. The results are shown (in milliseconds) in the box plot in Figure~\ref{fig:constr_times}. As can be seen, these computation times are generally far below one millisecond. For only a small number of solutions when $n=5000$, the solution evaluation times slightly exceed one millisecond. \\

Finally, the \texttt{STNWeb} tool~\citep{CHACONSARTORI2023100558} was employed to investigate the behavior of BRKGA and ACO when applied to instances of string length $n=5000$ and with all considered alphabet sizes. This tool generates visualizations, known as STN graphics, which depict the trajectories of algorithms as they explore the search space. We executed BRKGA and ACO 10 times on the first (out of 30) problem instances to create these trajectories, recording every new best-so-far solution. Figure~\ref{fig:STNWeb-graphics} presents the resulting STN graphics. The following observations can be made. First, when the alphabet is rather small ($|\Sigma| \in \{2, 4\}$, several different best solutions (red dots) are found by the algorithms. Moreover, in the case of $|\Sigma|=2$, both BRKGA and ACO find best solutions. Second, the trajectories of ACO seem slightly longer than those of BRKGA in the case of $|\Sigma|=2$, they seem more or less of the same length in cases $|\Sigma| \in \{4, 8\}$, and they become much shorter in the cases of larger alphabet sizes, that is, $|\Sigma|\in \{15, 32\}$. This indicates the growing difficulties of ACO with increasing alphabet size. Third, while the trajectory end points of ACO (grey triangles) are still close to the best solutions found by BRKGA in cases $|\Sigma|\in \{4, 8\}$, they are rather far from them for the largest two alphabet sizes. In addition, we would like to point out the large light-grey dots in Figures~\ref{fig:STNWeb-graphics:4} and~\ref{fig:STNWeb-graphics:5}, which indicate a confined area in the search space through which search trajectories from different algorithms have passed. Note that mostly search trajectories of ACO pass through these attraction points. However, while BRKGA seems able to escape from these areas, the ACO trajectories soon end after passing through these points. This indicates ACO's inability to escape from local minima.

\section{Conclusion}

We have presented a Biased Random Key Genetic Algorithm for solving a rather new problem from bio-informatics, the longest run subsequence problem, which is an NP-hard combinatorial optimization problem. For the design and implementation of our algorithm, we have paid special attention to the computational efficiency of the procedure for evaluating individuals. As was shown in the experimental result section, the time requirement of this procedure is generally way below one millisecond. A Max-Min Ant System for subset selection problems from the literature was implemented for the same problem for comparison purposes. The solution construction procedure of this algorithm also took profit from the time-efficient evaluation procedure originally developed for the Biased Random Key Genetic Algorithm. In addition, our algorithm was compared to the integer linear programming solver CPLEX. Experimental results for a diverse benchmark set of instances characterized by different input string lengths and alphabet sizes have shown that the Biased Random Key Genetic Algorithm outperforms the two competitors with statistical significance. The Max-Min Ant System can only compete for problem instances with very small alphabet sizes. \\

Future work will especially follow the following line. As shown by the experimental results, CPLEX produced slightly better solutions than our algorithm in the context of the shortest input strings with large alphabet sizes. This shows that there is room for improvement. We believe that such an improvement might be achieved by incorporating well-working heuristic information into the process of evaluating individuals. However, such heuristic information does currently not exist and requires some research effort to be developed.


\section*{Acknowledgments}

P.~Pinacho-Davidson acknowledges financial support from FONDECYT through grant number 11230359. C.~Blum was supported by grant PID2022-136787NB-I00 funded by MCIN/AEI/10.13039/ 501100011033.




\end{document}